\documentclass[letterpaper]{article} 
\usepackage{aaai25}  
\usepackage{times}  
\usepackage{helvet}  
\usepackage{courier}  
\usepackage[hyphens]{url}  
\usepackage{graphicx} 
\def\UrlFont{\rm}  
\usepackage{natbib}  
\usepackage{caption} 

\usepackage{amsmath} 
\usepackage{booktabs}
\usepackage{amssymb}
\usepackage[switch]{lineno}

\usepackage{xcolor}

\usepackage{algorithm}
\usepackage{algorithmic}

\usepackage{newfloat}
\usepackage{listings}
\DeclareCaptionStyle{ruled}{labelfont=normalfont,labelsep=colon,strut=off} 
\floatstyle{ruled}
\newfloat{listing}{tb}{lst}{}
\floatname{listing}{Listing}
\title{TraDiffusion: Trajectory-Based Training-Free Image Generation}
\author{
    Mingrui Wu\textsuperscript{\rm 1}\thanks{Equal Contribution.}, 
    	Oucheng Huang\textsuperscript{\rm 1$*$},
    	Jiayi Ji\textsuperscript{\rm 1},
     Jiale Li\textsuperscript{\rm 1},
     Xinyue Cai\textsuperscript{\rm 1},\\
     Huafeng Kuang\textsuperscript{\rm 1},
     Jianzhuang Liu\textsuperscript{\rm 2},
    	Xiaoshuai Sun\textsuperscript{\rm 1},
        Rongrong Ji\textsuperscript{\rm 1}\\ 
}
\affiliations{
    \textsuperscript{\rm 1}{Key Laboratory of Multimedia Trusted Perception and Efficient Computing, Ministry of Education of China,\\ Xiamen University, 361005, P.R. China}\\
    \textsuperscript{\rm 2}{Shenzhen Institute of Advanced Technology,
University of Chinese Academy of Sciences}

}

\usepackage{bibentry}

\begin{document}
\maketitle

\begin{abstract}
In this work, we propose a training-free, trajectory-based controllable T2I approach, termed TraDiffusion. This novel method allows users to effortlessly guide image generation via mouse trajectories.  To achieve precise control, we design a distance awareness energy function to effectively guide latent variables, ensuring that the focus of generation is  within the areas defined by the trajectory. The energy function encompasses a control function to draw the generation closer to the specified trajectory and a movement function to diminish activity in areas distant from the trajectory. Through extensive experiments and qualitative assessments on the COCO dataset, 
the results reveal that TraDiffusion facilitates simpler, more natural image control. Moreover, it showcases the ability to manipulate salient regions, attributes, and relationships within the generated images, alongside visual input based on arbitrary or enhanced trajectories. The code: \url{https://github.com/och-mac/TraDiffusion}.
\end{abstract}
%
\section{Introduction}
\begin{figure*}[t]
  \centering
  \includegraphics[width=0.95\linewidth]{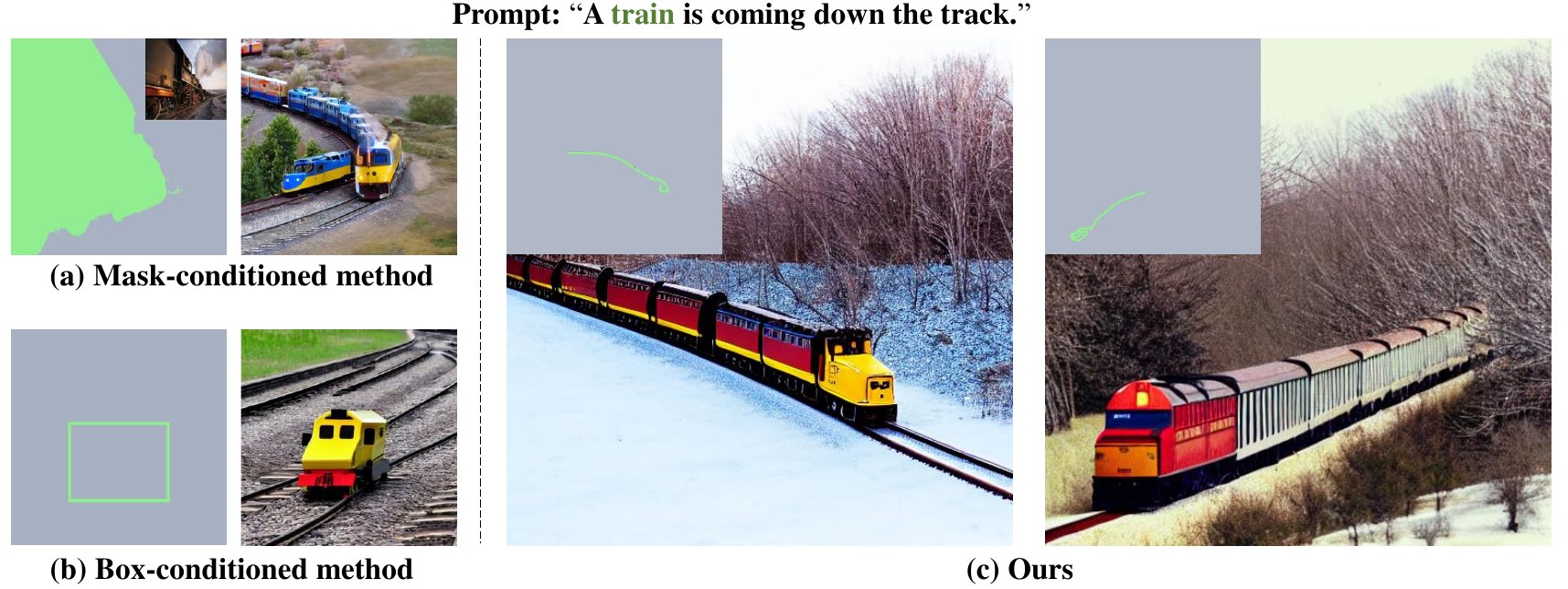}
  \caption{Comparing the mask-conditioned method~(a), box-conidtioned method~(b) and our trajectory-conditioned method~(c). The mask-conditioned method tends to have precise object shape control with a fine mask, which needs to be obtained by a specialized tool. The box-conidtioned methods enable coarse layout control. However, our trajectory-conditioned method provides a level of control granularity between the fine mask and the coarse box, which is user-friendly.}
  \label{fig:intro}
\end{figure*}

Over the past few years, the field of image generation has experienced remarkable progress, particularly with the development of models~\cite{goodfellow2020generative, ho2020denoising, rombach2022high, saharia2022photorealistic,ramesh2022hierarchical} trained on large-scale datasets sourced from the web. These models, particularly those that are text conditioned, have shown impressive capabilities in creating high-quality images that align with the text descriptions provided~\cite{dhariwal2021diffusion, song2020denoising, isola2017image, song2020score}. However, while text-based control has been beneficial, it often lacks the precision and intuitive manipulation needed for fine-grained adjustments in the generated images. As a result, there has been growing interest in exploring alternative conditioning methods~\cite{li2023gligen, nichol2021glide, zhang2020text, zhang2023adding}, such as edges, normal maps, and semantic layouts, to offer more nuanced control over the generated outputs. These diverse conditioning techniques broaden the scope of applications for generative models, extending from design tasks to data generation, among others.

Traditional methods~\cite{zhang2023adding, kim2023dense} with conditions such as edges, normal maps, and semantic layouts can achieve precise object shape control, while box-based methods enable coarse layout control. However, we find that trajectory-based control aligns more closely with actual human attention~\cite{xu2023pixel,pont2020connecting}, and provides a level of control granularity between the fine mask and the coarse box, as shown in Figure~\ref{fig:intro}.
Therefore, in parallel with these traditional layout control methods, this paper proposes a trajectory-based approach for text-to-image generation to fill this gap.


The central challenge we address is the utilization of trajectory to control image generation. Several studies~\cite{hertz2022prompt, kim2023dense, chen2024training} have successfully manipulated images by adjusting attention maps in the text-related cross-attention layers on the stable diffusion models~\cite{rombach2022high}, achieving effective control without additional training—a notably convenient approach. A standout method~\cite{chen2024training} among these, known as backward guidance, indirectly adjusts the attention by updating the latent variable. This technique, compared to direct attention map manipulation, yields images that are smoother and more accurately aligned with intended outcomes. It capitalizes on the straightforward nature of box-based conditioning, which effectively focuses attention within a specified bounding box region and minimizes it outside, enhancing the relevance of generated content. However, given the inherently sparse nature of trajectory-based control, applying backward guidance in this context poses significant challenges, requiring innovative adaptations to harness its potential effectively.

In this paper, we propose a novel training-free trajectory-conditioned image generation method. This technique enables users to guide the positions of image elements described in text prompts through trajectories, significantly enhancing the user experience by providing a straightforward way to control the appearance of generated images. To enable effective trajectory-based control, we introduce a distance awareness energy function. which updates latent variables, guiding the target to exhibit a stronger response in regions closer to the specified trajectory. The energy function comprises two main components: a control function, which directs the target towards the trajectory, and a movement function, which reduces the response in irrelevant areas distant from the trajectory.

Our trajectory-based approach offers a promising solution for layout-controlled image generation. Via qualitative and quantitative evaluations, we demonstrate the superior control capabilities of our method, achieving remarkable improvements in both the quality and accuracy of generated images. Moreover, our method exhibits adaptability to arbitrary trajectory inputs, allowing for precise control over object attributes, relationships, and salient regions. 


    
    

\section{Related Work}

\subsection{Image Diffusion Models}
Image diffusion models represent a pivotal advancement in the domain of text-to-image generation. These models~\cite{ho2020denoising, sohl2015deep,song2020score,avrahami2022blended,liu2022compositional,ruiz2023dreambooth, huang2024diffusion} operate by learning the intricate process of transforming textual descriptions into coherent and visually appealing images. One prominent approach within this paradigm is the Stable Diffusion Model (SDM)~\cite{rombach2022high}, which enhances the fidelity and stability of image generation.
The SDM is distinguished by its iterative denoising process initiated from a random noise map. This method, often performing in the latent space of a Variational AutoEncoder (VAE)~\cite{kingma2013auto, van2017neural}, enables the generation of images that faithfully captures the semantics conveyed in the input text. Notably, SDMs leverage pretrained language models~\cite{radford2021learning} to encode textual inputs into latent feature vectors, facilitating efficient exploration of the image manifold.
While image diffusion models excel in synthesizing images from textual prompts, accurately conveying all details of the image remains a challenge, particularly with longer prompts or atypical scenes. To address this issue, recent studies have explored the effectiveness of classifier-free guidance~\cite{ho2022classifier}. This innovative approach enhances the faithfulness of image generations by providing more precise control over the output, thereby improving the alignment with the input prompt.

\subsection{Controlling Image Generation with Layouts}
Layout controlled image generation introduces spatial conditioning to guide the image generation process. A lot of methods~\cite{feng2024layoutgpt,gafni2022make,hertz2022prompt,isola2017image,li2023gligen,liu2017unsupervised,wang2018high,xu2018attngan,zhang2023adding,zhang2021ufc,zhu2017unpaired,chen2024training,feng2022training,kim2023dense,xie2023boxdiff,yang2023reco,wang2024instancediffusion,bar2023multidiffusion,avrahami2023spatext,huang2023composer,huang2022multimodal,johnson2018image,park2019semantic,sun2019image,sylvain2021object,yang2022modeling,zhao2019image,qu2023layoutllm,li2021harmonious,tan2023alr,li2020bachgan,wu2022cross,qin2021layout,ren2024layered,zakraoui2021improving} offer different approaches to incorporate spatial controls for enhancing image synthesis.
GLIGEN~\cite{li2023gligen} and ControlNet~\cite{zhang2023adding} are notable examples that introduce finer-grained spatial control mechanisms. These methods leverage large pretrained diffusion models and allow users to specify spatial conditions such as Canny edges, Hough lines, user scribbles, human key points, segmentation maps, shape normals, depths, cartoon line drawings and bounding boxes to define desired image compositions. 
However, the advancement of spatially controlled image generation models have also brought significant training costs, stimulating the development of a range of training-free layout control and image editing methods~\cite{hertz2022prompt,xie2023boxdiff,kim2023dense}. These approaches leverage the inherent capabilities of cross-attention layers found in state-of-the-art diffusion models, which establish connections between word tokens and the spatial layouts of generated images. By exploiting this connection, these methods enable effective spatial control over the image synthesis process without the need for specialized training procedures.

\section{Preliminaries}
\subsection{Problem Definition}
We aim to improve layout control in image generation, which is formulated as $I = f(p, \{c_1, \cdots , c_n\})$, where the prompt $p$ and a set of layout conditions $\{c_1, \cdots , c_n\}$ are fed into the pretrained model $f$ to generate target image $I$. Given the model $f$, we hope to generate an image which aligns with the extra layout without further training or finetuning.

\subsection{Stable Diffusion}
Stable Diffusion (SD)~\cite{rombach2022high} is a modern text-to-image generator based on diffusion~\cite{saharia2022photorealistic}. SD consists of several key components: an image encoder and decoder, a text encoder, and a denoising network operating within a latent space.

During inference, the text encoder transforms the input prompt $p$ into a set of fixed-dimensional tokens $y = \{y_1, \cdots, y_m\}$. Then the denoising network, usually an UNet~\cite{ronneberger2015u} with cross-attention layers, takes a random noised sample latent code 
$z_t$
as input and returns $z_{t-1}$. This denoising process is iterated $t$ times to obtain the final latent code $z_0$. Finally, the latent code $z_0$ is fed into the image decoder to get the generated image.

In SD, the denoising network plays an important role in connecting the text condition and image information. Its core mechanism lies in the cross-attention layers. The cross-attention takes the transformed latent code $z^{(\tau)}$ in layer $\tau$ as query, and the transformed text conditions $y^{(\tau)}$ as keys and values, and the attention map is obtained as follows,
\begin{equation}
    A^{(\tau)}=\text{softmax}(\frac{z^{(\tau)} \cdot (y^{(\tau)})^T}{\sqrt{d_k}}),
\end{equation}
where $d_k$ is a scale factor, and $A^{(\tau)}$ consists of $A^{(\tau)}_i$, $i \in \{1, \cdots, m\}$, representing the impact of the $i$-th token on the output. 


\section{Method}
\begin{figure}[t]
\centering
  \centering
  \includegraphics[width=0.9\linewidth]{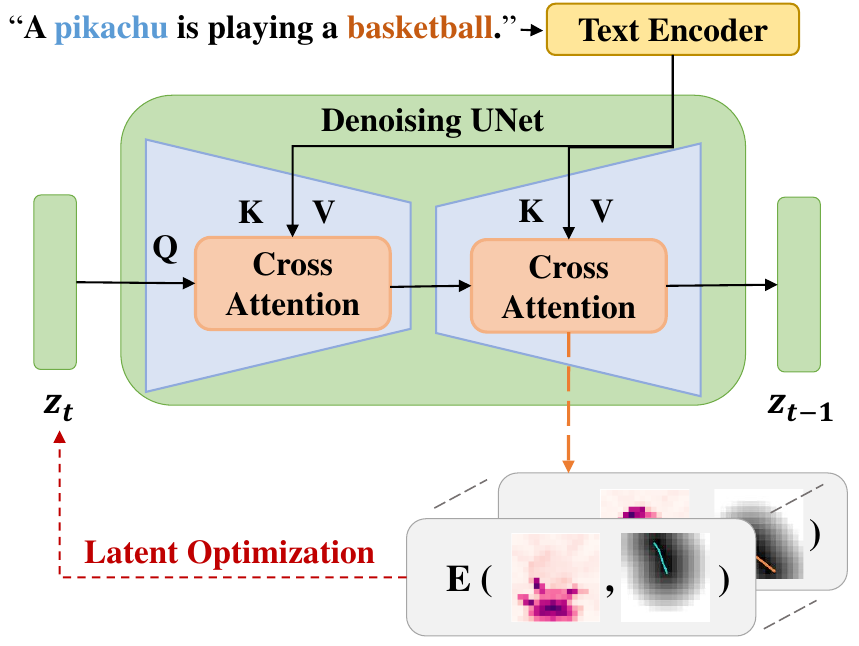}
  \caption{Overview of the distance awareness guidance. With the provided trajectories, we calculate distance matrices for each trajectory. Subsequently, we compute the distance awareness energy function between these distance matrices and the attention map of each object. Finally, during the inference process, we conduct backpropagation to optimize the latent code.}
  \vspace{-0.25cm}
  \label{fig:archi}
\end{figure}

In this section, we introduce the trajectory-based controllable text-to-image generation method (as shown in Figure~\ref{fig:archi}) using the pretrained diffusion model~\cite{rombach2022high}, and describe the distance awareness energy function that combines the trajectory to achieve training-free layout control.

\subsection{Controlling Image Generation with Trajectory}
Previous works~\cite{kim2023dense, xie2023boxdiff, chen2024training} are mainly based on masks or boxes to control the layout, but masks are fine-grained, which is not user-friendly, and boxes are too coarse to limit the object area. These methods directly affect the prior structure of the generated object in the image. In some cases, we only want to guide the approximate location and shape of the object, rather than limiting the object to a specified shape or size. So we introduce trajectories to guide the layout of the generated image. Specifically, we provide a trajectory for a specified word or phrase in the prompt. The problem can be formulated as 
$I = f(p, \{(w_1, l_1), \cdots , (w_n, l_n)\})$,
where $p$ represents the global prompt, and a set of word-line pairs $(w_i, l_i)$ serving as layout conditions, which are fed into the pretrained model $f$ to generate the target image $I$.
Based on the trajectories, we guide the locations of instances, attributes, relationships and actions without further training or finetuning. And the user can easily draw trajectories for image generation through the mouse or pen.


\subsection{Distance Awareness Guidance}
Inspired by~\cite{chen2024training}, we try to control the image generation based on trajectories with backward guidance. However, due to the sparsity of the trajectories, it is difficult to directly combine backward guidance. A natural idea is to get the prior structure of an object through the attention maps of cross-attention layers, rather than directly using the trajectories to achieve backward guidance.

\begin{figure}[t]
  \centering
  \includegraphics[width=0.93\linewidth]{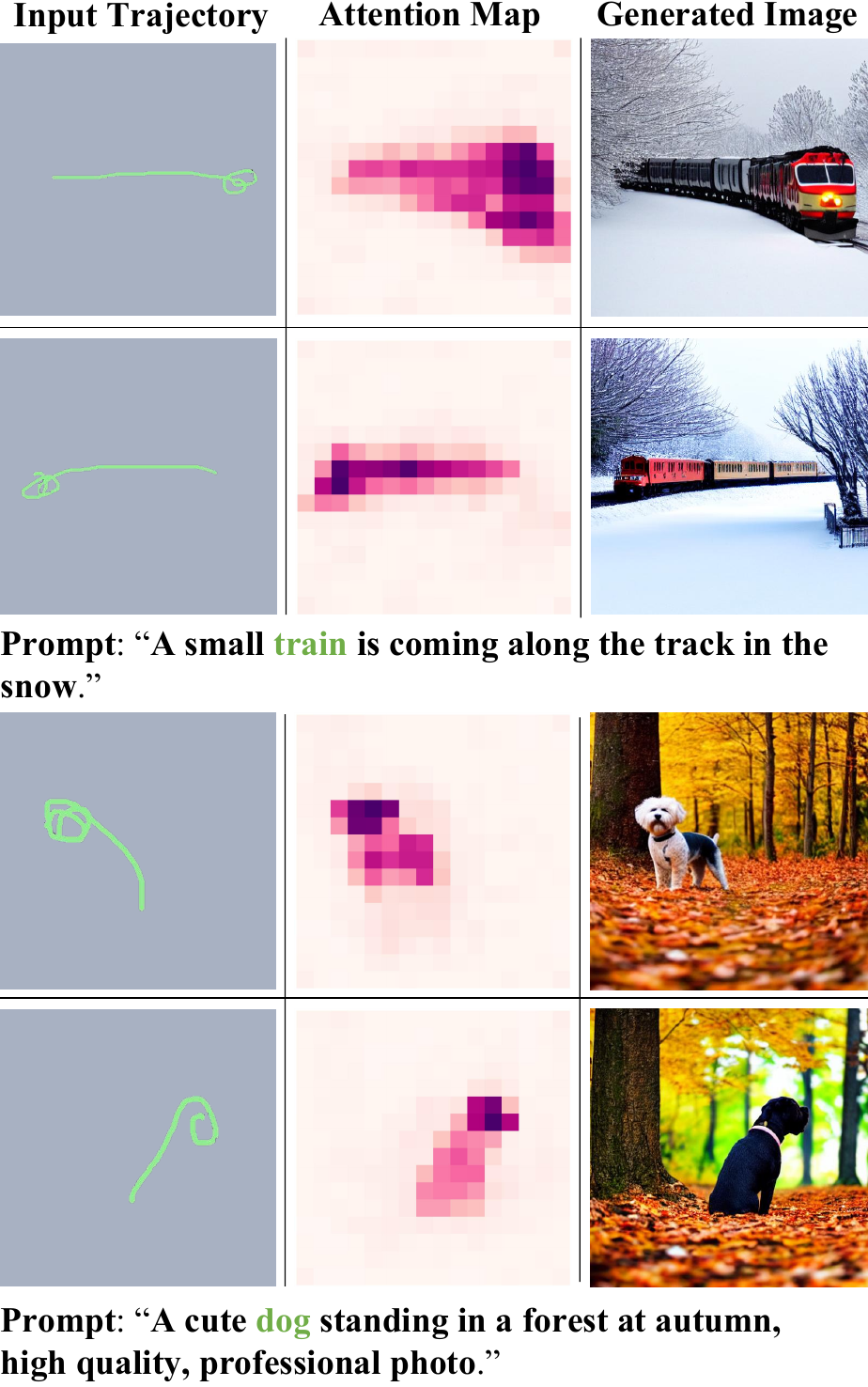}
  \caption{Examples of controlling the salient areas of the objects with trajectories. We can adjust the position of the local salient area of the object by enhancing the local trajectory.}
  \label{fig:w_t}
\end{figure}

\begin{figure}[t]
  \centering
  \includegraphics[width=0.93\linewidth]{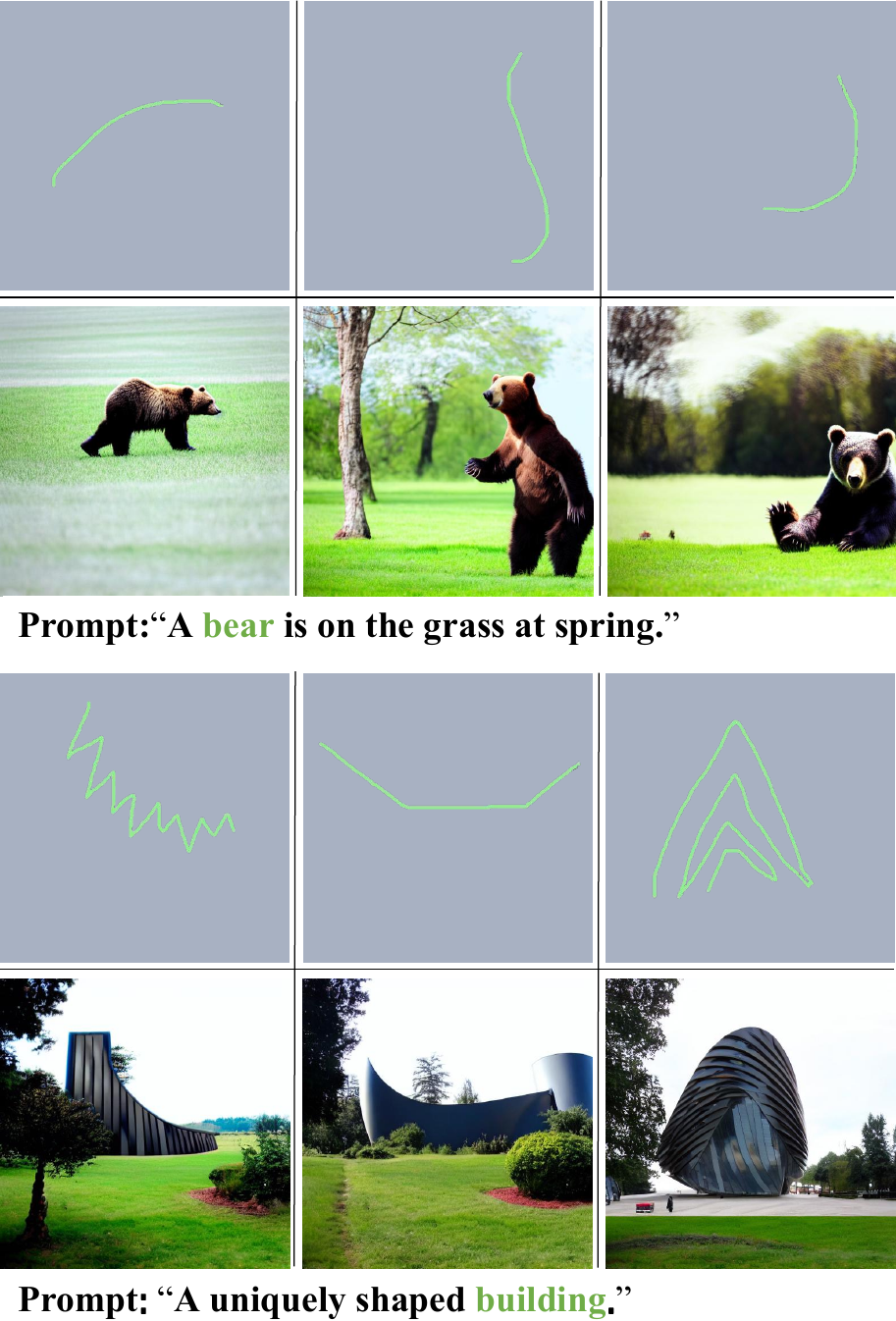}
  \caption{Examples of controlling the object shapes with arbitrary trajectories. We can adjust the posture of the object~(top) or specify the approximate shape of the object~(bottom) by varying the given trajectory.}
  \label{fig:arbit}
\end{figure}

\subsubsection{Prior Structure Based Guidance.}
To get the prior structure of an object, we first perform denoising of the $T_k$ steps on the Stable Diffusion model and apply a threshold on the attention map of the current step to obtain a binary mask. Then we move the mask to align the center of the trajectory. By this, we can use this mask to replace the box to compute the energy function proposed in~\cite{chen2024training}.

However, we find that this approach has several unavoidable drawbacks, as shown in Figure~\ref{fig:mask_err} of Appendix. a) In order to get a good quality mask, we have to carefully select the appropriate threshold, as well as suitable denoising steps. Too many denoising steps would produce a fine mask but at the same time introduce an excessive amount of additional computation and an overfitting object prior. b) Since the Stable Diffusion model does not always produce high-quality images, it always produces some unusable masks in some cases. Taken together, prior structure based guidance cannot be a robust guidance strategy.

\subsubsection{Distance Awareness Energy Function.}
To overcome the above limitations of prior structure based guidance, we propose to use a distance awareness energy function for guidance, as shown in Figure~\ref{fig:archi}. Specifically, we first apply a control function to guide the object to approach a given trajectory, which is formulated as
\begin{equation}
    E_c\left(A^{(\tau)}, l_i, w_i\right)=(1-\frac{\sum_\mu (D_{\mu i}+\epsilon)^{-1} A_{\mu i}^{(\tau)} }{\sum_\mu A_{\mu i}^{(\tau)}})^2,
\end{equation}
where $D_{\mu i}$ is a distance matrix computed by the OpenCV~\cite{bradski2000opencv} function ``$distanceTransform$'', in which each value denotes the distance from each location $\mu$ of the attention map to the given trajectory $l_i$, $\epsilon$ is a very small value used to avoid division by zero, and $A^{(\tau)}_{\mu i}$ is the attention map determining how strongly each location $\mu$ in layer
$\tau$ is associated with the $i$-th token $w_i$. This function steers the object to approach the given trajectory. 

However, this does not effectively inhibit the attention response of the object in irrelevant regions far from the trajectory. So, we add a movement function to suppress the attention response from irrelevant regions far from the trajectory of the object accordingly. The movement function is formulated as
\begin{equation}
    E_m\left(A^{(\tau)}, l_i, w_i\right)=(1-\frac{\sum_\mu A_{\mu i}^{(\tau)}}{\sum_\mu D_{\mu i} A_{\mu i}^{(\tau)} })^2.
\end{equation}

The final distance awareness energy function is the combination of $E_c$ and $E_m$:
\begin{equation}
    E=E_c + \lambda E_m,
\end{equation}
where $\lambda$ is an adjustable hyperparameter. By computing $E$ as loss and backpropagation to update the latent $z_t$, we encourage the response of the cross-attention map of the $i$-th token to obtain higher values in the area close to the trajectory $l_i$, which can be formulated as
\begin{equation}
\boldsymbol{z}_t \leftarrow \boldsymbol{z}_t-\sigma_t^2 \eta \nabla_{\boldsymbol{z}_t} \sum_{\tau \in \Phi} \sum_{i \in \mathbb{N}} E\left(A^{(\tau)}, l_i, w_i\right),
\end{equation}
where $\eta>0$ is a hyperparameter controlling the strength of the guidance, $\Phi$ is a set of layers in UNet~\cite{ronneberger2015u}, $\mathbb{N}=\{1, \cdots , n\}$, and $\sigma_t=\sqrt{\left(1-\alpha_t\right) / \alpha_t}$, with $\alpha_t$ being a pre-defined parameter of diffusion~\cite{ho2020denoising,rombach2022high,song2020denoising}.

\begin{figure}[t]
  \centering
  \includegraphics[width=0.93\linewidth]{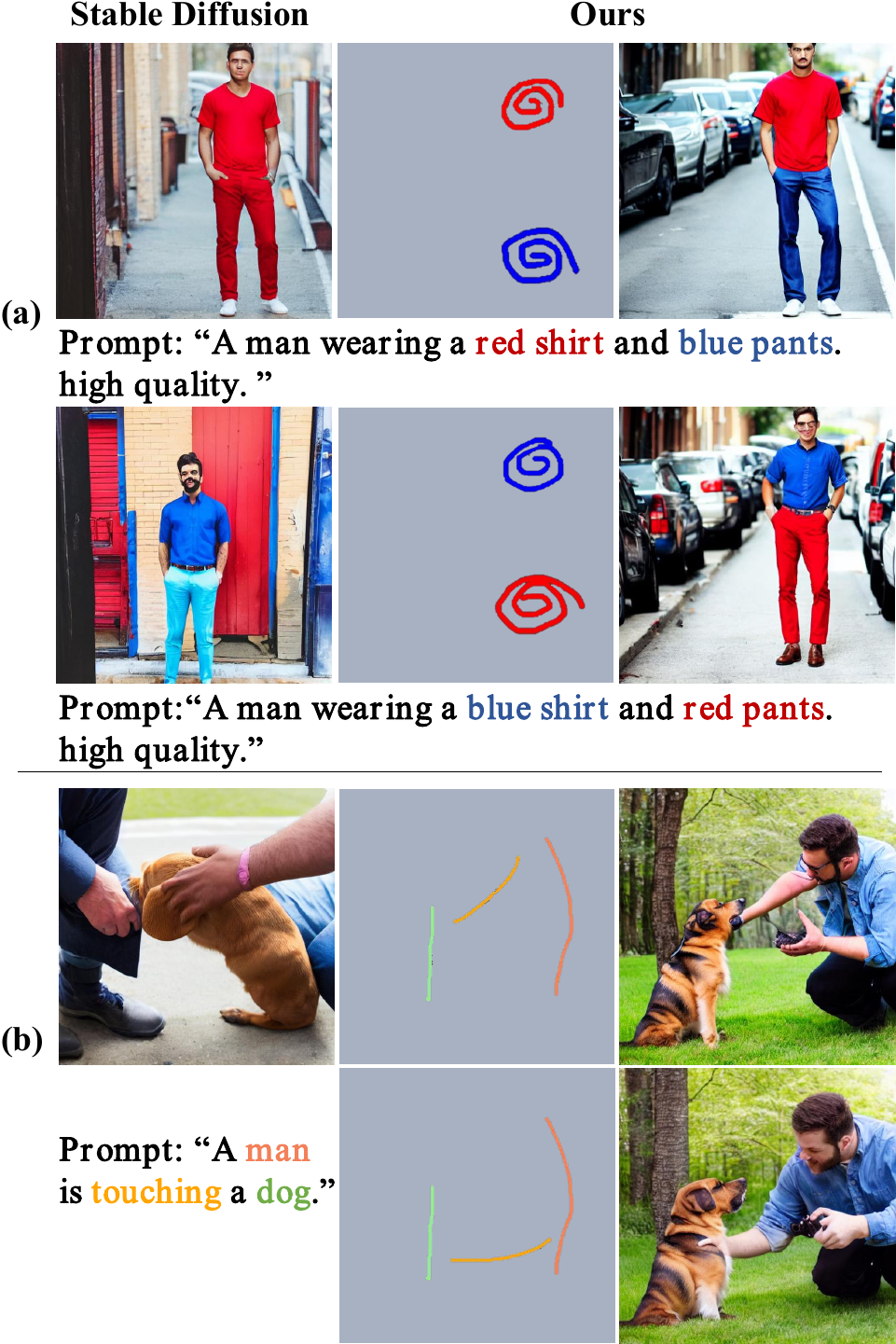}
  \caption{Examples of controlling the attribute and relationship of objects. Based on trajectories, we can overcome the attribute confusion issue of the pre-trained Stable Diffusion model, generating visual results consistent with the given prompt~(a), and adjust the positions of interactions~(b).}
  \label{fig:rel_attr}
\end{figure}

\begin{figure}[t]
  \centering
  \includegraphics[width=0.93\linewidth]{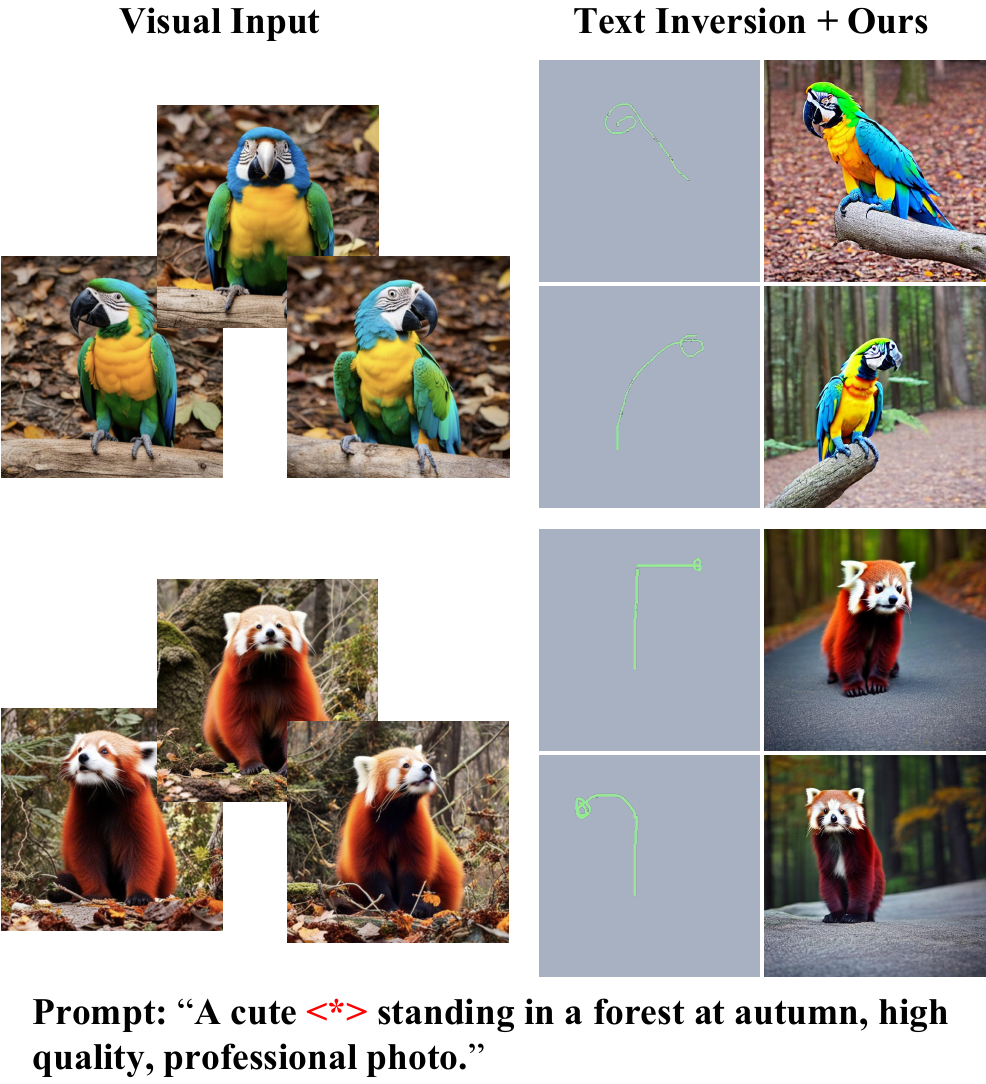}
  \caption{Examples of controlling visual input. }
  \label{fig:text_v}
\end{figure}

\begin{figure*}[t]
  \centering
  \includegraphics[width=0.99\linewidth]{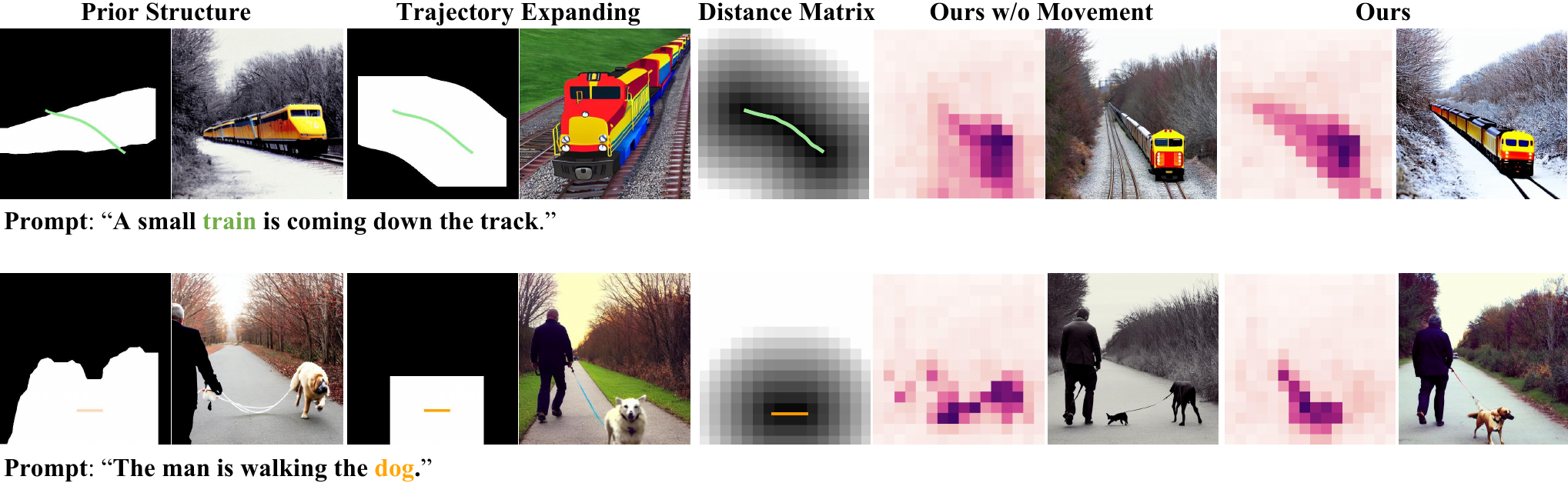}
  \caption{Qualitative analysis of the components in our proposed method, including prior structure based guidance~(left), expanding the trajectory to obtain a mask~(middle), and our method without and with the movement function~(right). We show the input condition and generated image for each component, and an extra attention map for our method.}
  \label{fig:abl}
\end{figure*}

\section{Experiments}
\subsection{Experimental Setup}
\subsubsection{Evaluation Benchmark.} 
We evaluate our approach on COCO2014~\cite{lin2014microsoft}. Following previous works~\cite{bar2023multidiffusion, chen2024training}, we randomly select 1000 images from its validation set, and each image is paired with a caption and has up to 3 instances with masks that occupy more than 5\% of the image.
However, the instances that are randomly sampled may not appear in the caption, so the previous works~\cite{bar2023multidiffusion, chen2024training} pad the instance names into the caption. But this inevitably changes the effect of the prompt on generating images, so we prioritize sampling images with instances in the captions rather than padding the captions.

\subsubsection{Evaluation Metrics.} We measure the quality of the generated images with FID. 
However, the traditional metrics are not suitable for evaluating the layout control of trajectory-based image generation methods, so we propose a novel \textit{Distance To Line}~(DTL) metric, which is defined as
\begin{equation}
    DTL=\frac{1}{n} \sum_{i \in \mathbb{N}} \frac{\sum_{\mu \in mask} e^{-D_{\mu i}}}{\sum_{\mu \in mask} 1},
\end{equation}
where \textit{mask} is obtained by applying the YOLOv8m-Seg~\cite{yolov8_ultralytics, redmon2016you} on the generated image, and $\mathbb{N}=\{1, \cdots , n\}$. The larger the DTL, the closer the generated object is to the given trajectory. Therefore, DTL not only verifies whether the desired objects are generated but also examines the alignment of the layout. We report mean DTL on all generated images.

\subsubsection{Implementation Details.}
Following the setting of ~\cite{chen2024training}, we utilize Stable-Diffusion~(SD) V-1.5~\cite{rombach2022high} as the default pre-trained diffusion model. We select the cross-attention maps of the same layers as ~\cite{chen2024training} for computing the energy function. And the backpropagation of the energy function is performed during the initial 10 steps of the diffusion process and repeated 5 times at each step. The hyperparameters $\lambda=10$ and $\eta=30$. We fix the random seeds to 450. The experiments are performed on a RTX-3090 GPU.



\subsection{Applications}

\subsubsection{Controlling the Salient Areas of Objects.}
Typically, attention models exhibit higher responses in salient regions of objects~\cite{xu2015show, oktay2018attention, zhang2019self, zeiler2014visualizing, hu2018squeeze}. Hence, we investigate whether enhancing local trajectories can effectively control the positions of salient regions within objects. As illustrated in Figure~\ref{fig:w_t}, 
we showcase our method's capability to guide attention maps by manipulating local trajectories, thereby exerting control over the positioning of specific elements such as the train's head and the dog's head.

\subsubsection{Controlling Shapes with Arbitrary Trajectories.} 
We analyze the adaptability of our method to incorporate trajectory inputs of arbitrary shapes to generate the desired object shapes. As illustrated in Figure~\ref{fig:arbit}, by varying the trajectory, we can adjust the posture of the object, such as guiding the posture of a `bear' into various positions such as crawling, standing, and sitting~(Figure~\ref{fig:arbit} top). Additionally, we can specify the approximate shape of the object by the trajectory~(Figure~\ref{fig:arbit} bottom). 


\subsubsection{Controlling Attributes and Relationship.}
We analyze whether our method can control the attributes of objects and the relationships between objects. As illustrated in Figure~\ref{fig:rel_attr}, attribute confusion exists in the SD model. Despite our efforts to generate the shirts and pants in varied colors, it persistently confuses the attributes, resulting in the wrong colors for both. By controlling the attributes of the object based on trajectories, we can largely overcome the attribute confusion issue in the pre-trained Stable Diffusion model, generating visual results consistent with the given prompt~(Figure~\ref{fig:rel_attr} a). Additionally, we can adjust the positions of interactions between objects by adjusting the trajectories~(Figure~\ref{fig:rel_attr} b).

\subsubsection{Controlling Visual Input.}
We analyze whether our method can control the visual input. As shown in Figure~\ref{fig:text_v}, we can adjust the orientations of the visual input objects through trajectories. However, it is worth noting that finer adjustments pose challenges, which relies on the available visivility of the input objects.


\begin{table}[t]
  \centering
  \setlength{\tabcolsep}{1mm}
  \begin{tabular}{lcc}
    \toprule
    Method & DTL($\uparrow$) & FID($\downarrow$)\\
    \midrule
    Stable Diffusion~\cite{rombach2022high} & 0.0043 & 68.33  \\
    Prior structure & 0.0077 & 66.91 \\
    Trajectory expanding & 0.0080 & 64.87 \\
    \midrule
    Ours w/o movement & 0.0119 & 64.68 \\
    Ours  & 0.0156 & 68.53 \\
    \bottomrule
  \end{tabular}
  \caption{Ablation study on each component of our method. Compared to the prior structure based guidance method and the trajectory expanding method, our method demonstrates the strongest level of control, with a DTL score about twice as high as those of the two baselines.}
  \label{tab:abl}
\end{table}

\begin{table*}[t]
    \centering
\scalebox{0.92}{ \begin{tabular}{lc|ccc}
    \toprule
    Method & Type & Quality($\uparrow$) & Controllability($\uparrow$) & User-Friendliness($\uparrow$)  \\
    \midrule
    BoxDiff & Box & 3.52 & 3.22 & 2.07  \\
    Backward Guidance & Box & 3.24 & 3.69 &  2.07  \\
    DenseDiffusion & Mask & 3.30 & 3.56 & 1.07   \\
    \midrule
    Ours & Trajectory & 3.72 & 4.04 & 2.87   \\
    \bottomrule
  \end{tabular}
  }
  \caption{The user studies, including quality, controllability (score from 1 to 5), and user-friendliness (score from 1 to 3).}
  \label{tab:sota}
\end{table*}

\subsection{Ablation Study}
We perform the ablation study to validate the effect of each component in our proposed method. We first evaluate the Stable Diffusion model~\cite{rombach2022high} for reference. We consider the prior structure based guidance as the baseline, and a method of expanding to the fixed size outwards along the trajectory to obtain a mask is also compared. Then we experiment with only the control function to validate the controllability, and further add the movement function to verify that the method is able to suppress the response of object at the irrelevant regions far from the trajectory.

The results are shown in Table~\ref{tab:abl}. We can observe that the prior structure based guidance and the trajectory expanding methods exhibit similarly low DTL scores. However, our method shows an improvement of about 50\% in DTL compared to the two baselines when the movement loss is not added. Through further augmentation by the movement loss, our method demonstrates a significant 100\% enhancement in DTL.  
Although there is a slight decrease in FID performance after adding the movement loss, we believe that this minor difference can be negligible due to the complexity of the COCO image distribution.

The qualitative analysis of the components in our proposed method is shown in Figure~\ref{fig:abl}. We can observe that both of the trajectory expanding based method and the prior structure based guidance method fail to generate outputs that strictly adhere to the trajectory control, potentially resulting in similar issues encountered with the box-based and mask-based approaches. Additionally, mask-based methods may struggle to capture effective prior structures of the objects. In contrast, our approach, without introducing additional movement loss, is capable of generating objects that adhere to the trajectory~(top). However, due to the lack of suppression of irrelevant positions in the attention far away from the given trajectory, extra object generations occur~(bottom). This issue is alleviated by further adding the movement loss.


The effect of the hyperparameter $\lambda$ is shown in Table~\ref{tab:hyper} and Figure~\ref{fig:hyper} of Appendix. It shows that when $\lambda=20$, it yields the highest DTL results. However, we also notice a comparable performance when $\lambda=10$, and increasing $\lambda$ further leads to a significant decrease in FID. In addition, as shown in Figure~\ref{fig:hyper} of Appendix, we observe that excessively large values lead to over-suppression of the entire image, while values in the range of [5,10] yield the best results. Therefore, the default $\lambda$ is set to 10.


\subsection{Comparison with Prior Work}
We compare our method with previous layout text-to-image generation methods, including mask-conditioned DenseDiffusion~\cite{kim2023dense} and ControlNet~\cite{zhang2023adding}, and box-conditioned BoxDiff~\cite{xie2023boxdiff} and Backward Guidance~\cite{chen2024training}, in which DenseDiffusion, BoxDiff, and Backward Guidance are all training-free.  
In our method, we sample the trajectories inside the boxes or masks.
Typically, existing evaluation metrics, like YOLO-score and mIOU, are inevitably biased towards each type of layout control method due to the lack of a unified and feasible metric for comparison. To address this, we compare our method with previous training-free methods by providing user studies on the results' quality, controllability, and user-friendliness, based on the average scores from 15 users, as shown in Table~\ref{tab:sota}. 

The visual examples of the comparisons are shown in Figure~\ref{fig:vis1} of Appendix. Mask-based methods often introduce excessive manual priors by utilizing too detailed masks, leading to the overly controlled generation of distorted and unrealistic objects. For example, this can be observed in the generation of the distorted airplanes~(c) and elephants~(d). Conversely, box-based methods, with their too coarse control conditions, completely disregard prior information about the object, leading to the generation of deformed and unnatural images, such as the floating frisbee~(a), oversized umbrella~(b), and snowboard depicted at a unreasonable angle~(e). In contrast, our trajectory-based approach does not excessively intervene in the prior structure of the object and, with user-friendly simple controls, is capable of generating natural images. 

In addition, it is noteworthy that trained layout text-to-image generation methods often have limitations in accommodating diverse semantic categories and conditional domains. This often necessitates retraining to adapt to new conditions, incurring additional cost and time.
However, our innovative training-free method can seamlessly adapts the model to any semantic input,
offering unparalleled convenience and flexibility to users.


\section{Limitations}
While we have demonstrated simple and natural layout control by  trajectory, our method is subject to a few limitations. 
Firstly, same as other training-free layout control text-to-image generation methods, the quality of images generated based on trajectory is limited by the pre-trained SD model. Adjustments to both the prompt and trajectory may be necessary to achieve desired outcome. 
Secondly, similar to ~\cite{chen2024training}, we also incur twice the inference cost compared to the pre-trained SD model.  
Thirdly, although trajectories are less coarse than bounding boxes, achieving precise adjustments to the shapes of objects remains challenging.
Fourthly, we have currently only explored a limited range of possibilities in trajectory-based image generation, and we look forward to further exploration of its diverse applications in future work. 

\section{Conclusions}
In this work, we propose a trajectory-based layout control method for text-to-image generation without additional training or fine-tuning. Combining with the proposed distance awareness energy function to optimize the latent code of the Stable Diffusion model, we achieve user-friendly layout control. In the energy function, the control function steers the object to approach the given trajectory, and the movement function inhibits the response of the object in irrelevant regions far from the trajectory. A set of experiments show that our method can generate images more simply and naturally. Moreover, it exhibits adaptability to arbitrary trajectory inputs, allowing for
precise control over object attributes, relationships, and salient
regions. We hope that our work can inspire the community to explore more user-friendly text-to-image techniques, as well as uncover more trajectory-based applications.

\section*{Acknowledgments}
This work was supported by National Science and Technology Major Project (No. 2022ZD0118201), the National Science Fund for Distinguished Young Scholars (No.62025603), the National Natural Science Foundation of China (No. U21B2037, No. U22B2051, No. 62072389, No. 62302411), China Postdoctoral Science Foundation (No. 2023M732948), the Natural Science Foundation of Fujian Province of China (No.2022J06001), and partially sponsored by CCF-NetEase ThunderFire Innovation Research Funding (NO. CCF-Netease 202301).

\bibliography{aaai25}

\newpage
\appendix
\section{Appendix}
\subsection{Comparison with Prior Work}
We compare our method with previous text-to-image generation methods with layout control on traditional metrics, including mask-conditioned methods DenseDiffusion, and box-conditioned methods BoxDiff and Backward Guidance, in which DenseDiffusion, BoxDiff and Backward Guidance are all training-free methods. 

\begin{table}[h]
  \centering
  \begin{tabular}{lcc}
    \toprule
    Method & FID($\downarrow$) & CLIP-score($\uparrow$)\\
    \midrule
    BoxDiff & 71.73 & 30.03 \\
    Backward Guidance & 69.04 & 30.76  \\
    DenseDiffusion & 74.70 & 30.34  \\
    \midrule
    Ours & 68.53 & 30.78   \\
    \bottomrule
  \end{tabular}
  \caption{Comparison with prior works on traditional metrics.}
  \label{tab:sota1}
\end{table}

\begin{figure*}[t]
  \centering
  \includegraphics[width=0.93\linewidth]{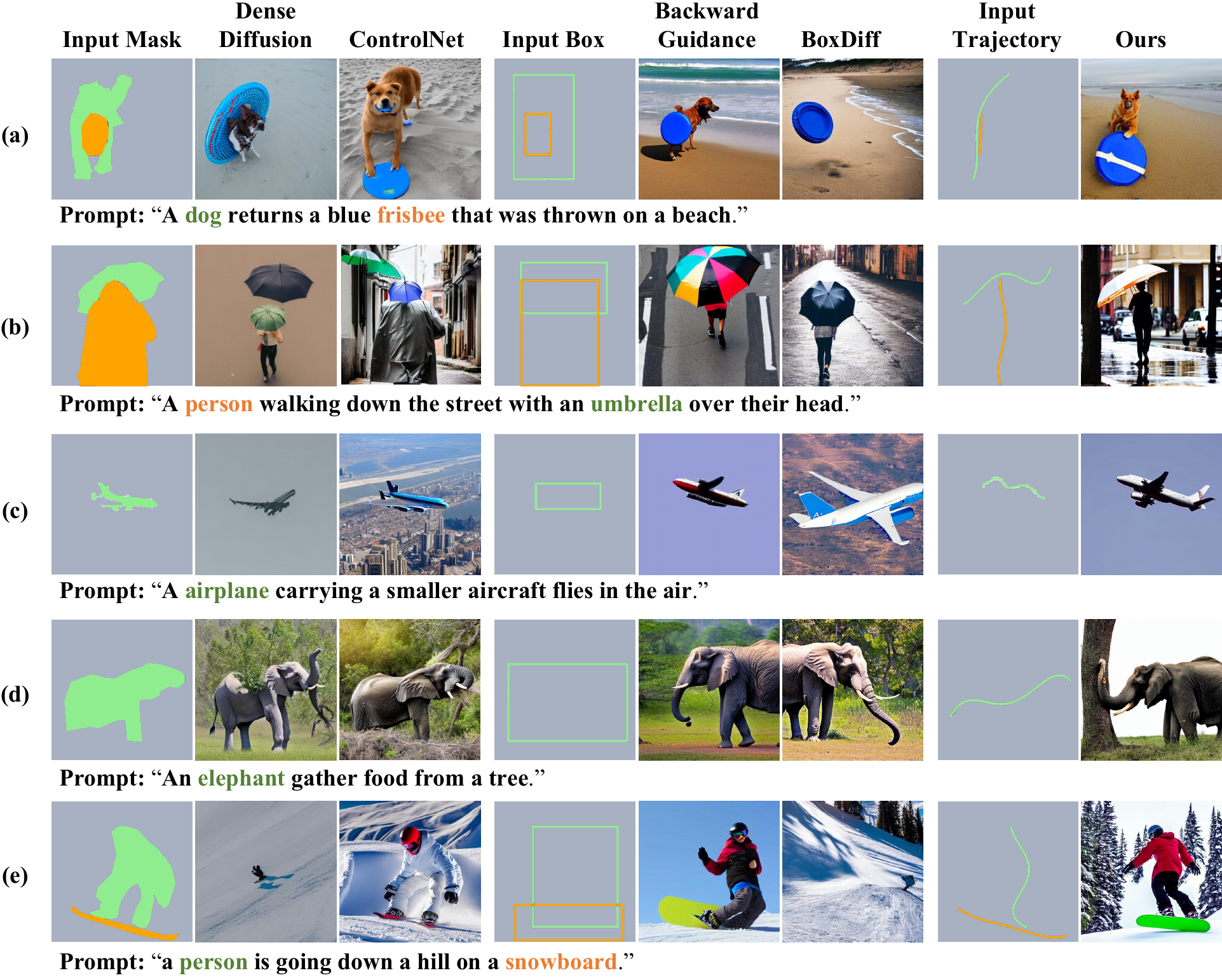}
  \caption{Qualitative comparison with prior mask-based and box-based layout control works. The controlled targets are colored with green and orange. The mask-based and box-based layout control methods generate the unnatural images due to the control conditions that are too fine or too coarse. However, our simple trajectory-based approach yields more natural results.}
  \label{fig:vis1}
\end{figure*}

\begin{figure*}[t]
  \centering
  \includegraphics[width=0.99\linewidth]{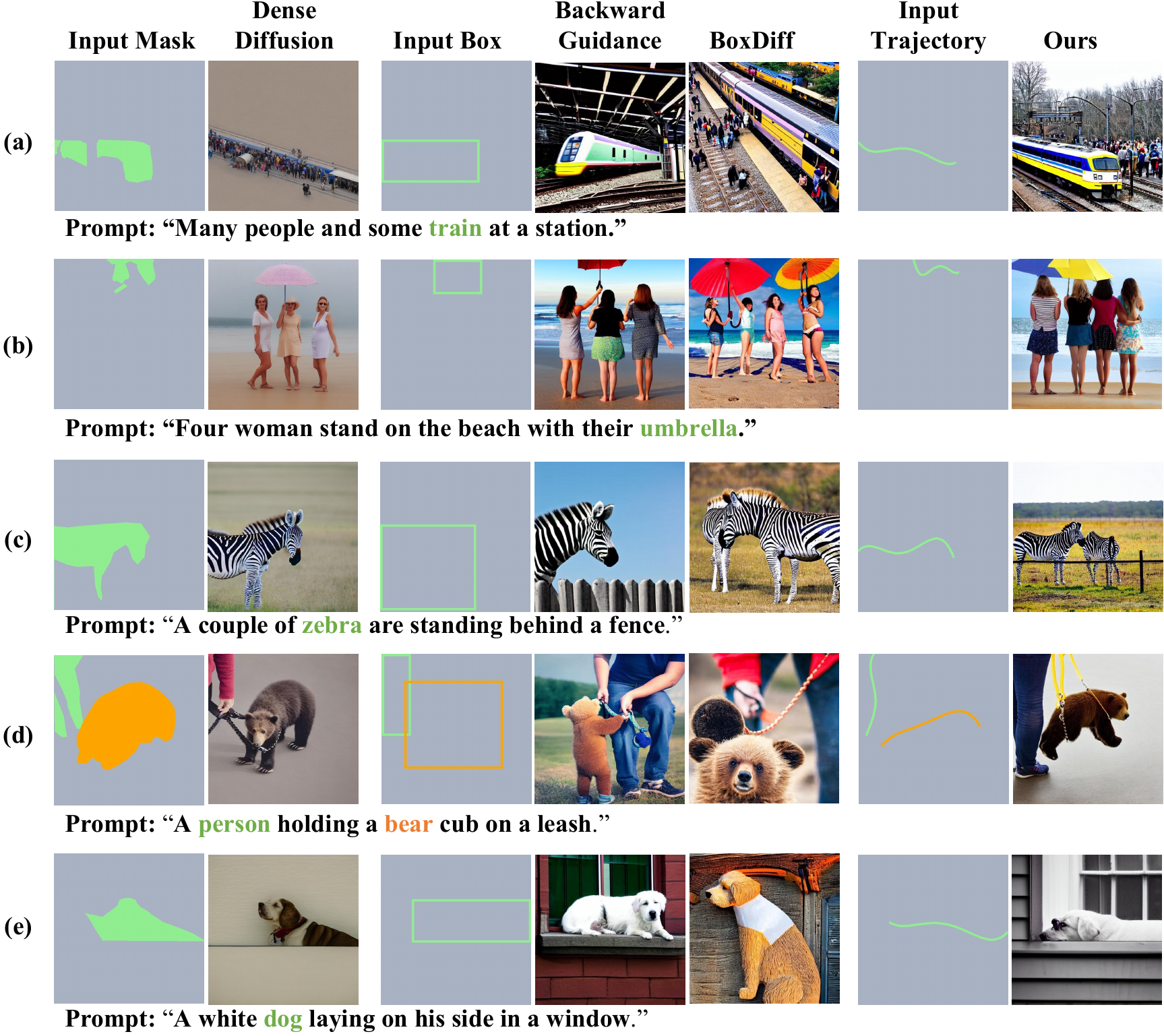}
  \caption{More examples of comparing with prior works.}
  \label{fig:vis2}
\end{figure*}

The examples as shown in Figure~\ref{fig:vis1}. In our implementation, ControlNet does not support the categories ``dog'', ``frisbee'', ``umbrella'', ``elephant'' and ``snowboard'', so we employee the superclass ``animal'' to replace ``dog'' and ``elephant'', and do not control the ``frisbee'', ``umbrella'' and ``snowboard''. In contrast, our training-free method can adapt to any semantic input.
And more examples as shown in Figure~\ref{fig:vis2}. We remove ControlNet in Figure~\ref{fig:vis2} due to it cannot support most of semantic categories.

\begin{table}[h]
  \vspace{0.1cm}
  \centering
  \setlength{\tabcolsep}{1mm}
  \begin{tabular}{lcccccc}
    \toprule
    $\lambda \rightarrow$  & 0 & 1 & 5 & 10 & 20 & 100 \\
    \midrule
    DTL($\uparrow$) & 0.0119 & 0.0124 & 0.0137 & 0.0156 & 0.0158 & 0.0096  \\
    FID($\downarrow$) & 64.68 & 65.46 & 66.39 & 68.53 & 72.80 & 129.91  \\
    \bottomrule
  \end{tabular}
   \caption{Ablation study on the effect of the hyperparameter $\lambda$. The best performance is achieved when $\lambda$ is around 10.}
    \label{tab:hyper}
\end{table}


\subsection{The Effect of Additional Conditions}
\begin{figure}[t]
  \centering
  \includegraphics[width=0.95\linewidth]{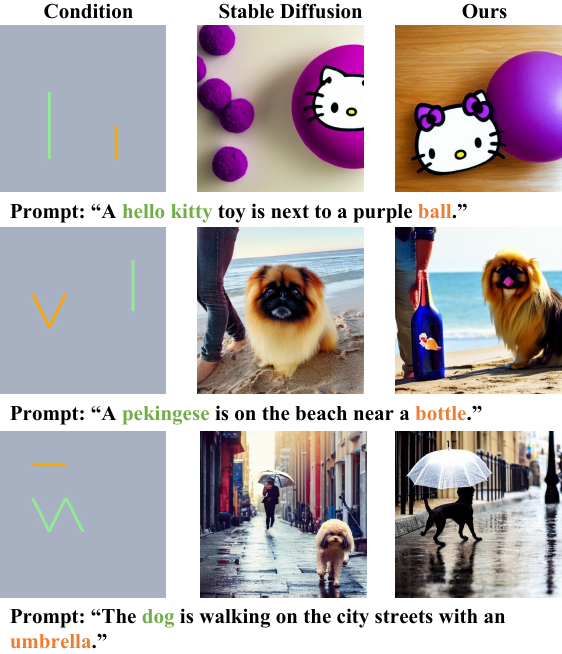}
  \caption{Comparing with pretrained Stable Diffusion model. Our method can guide Stable Diffusion model to generate multiple targets, despite the inherent limitations of the Stable Diffusion model in this regard.}
  \label{fig:sd_fail}
\end{figure}

\begin{figure}[!htb]
  \centering
  \includegraphics[width=0.95\linewidth]{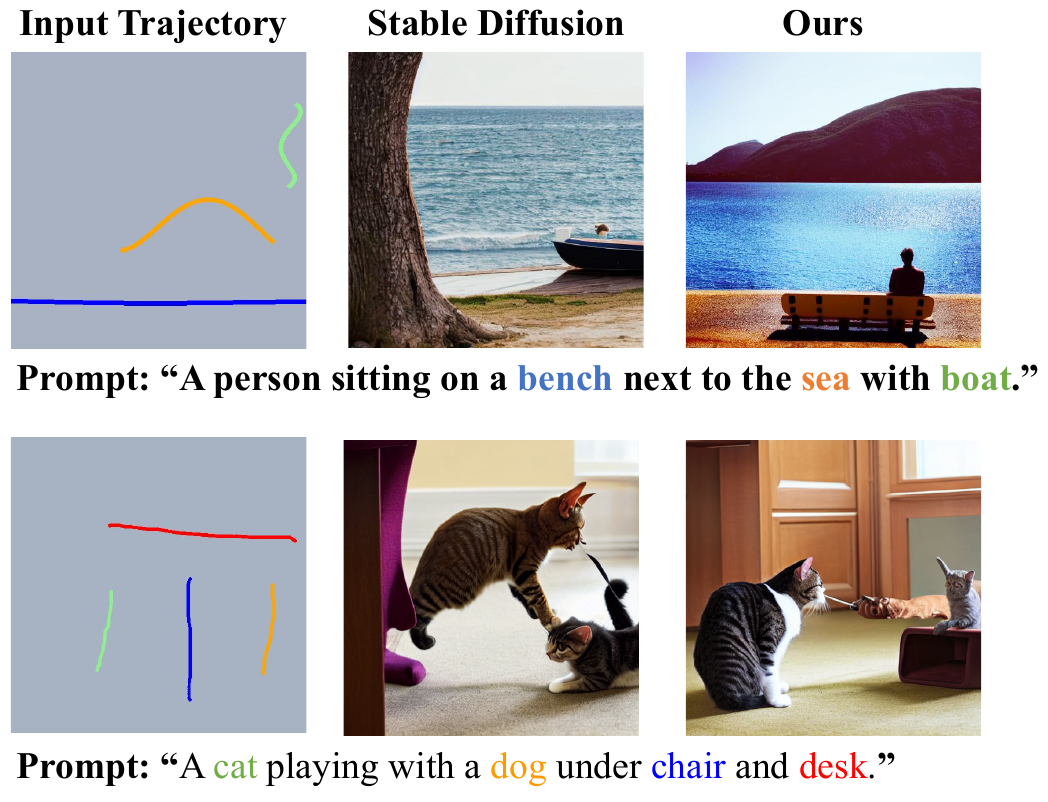}
  \caption{The examples of failed cases. Our approach fails in controlling more targets, which may be related to the intrinsic mechanism of the stable diffusion model.}
  \label{fig:fail}
\end{figure}

We compare our trajectory-based method with pretrained Stable Diffusion model, as shown in Figure~\ref{fig:sd_fail}, we observe that the Stable Diffusion model often struggles when generating multiple targets. However, by incorporating additional control conditions, our approach successfully achieves the intended targets. And the examples of failed cases as shown in Figure~\ref{fig:fail}.
\begin{figure}[H]
  \centering
  \includegraphics[width=0.95\linewidth]{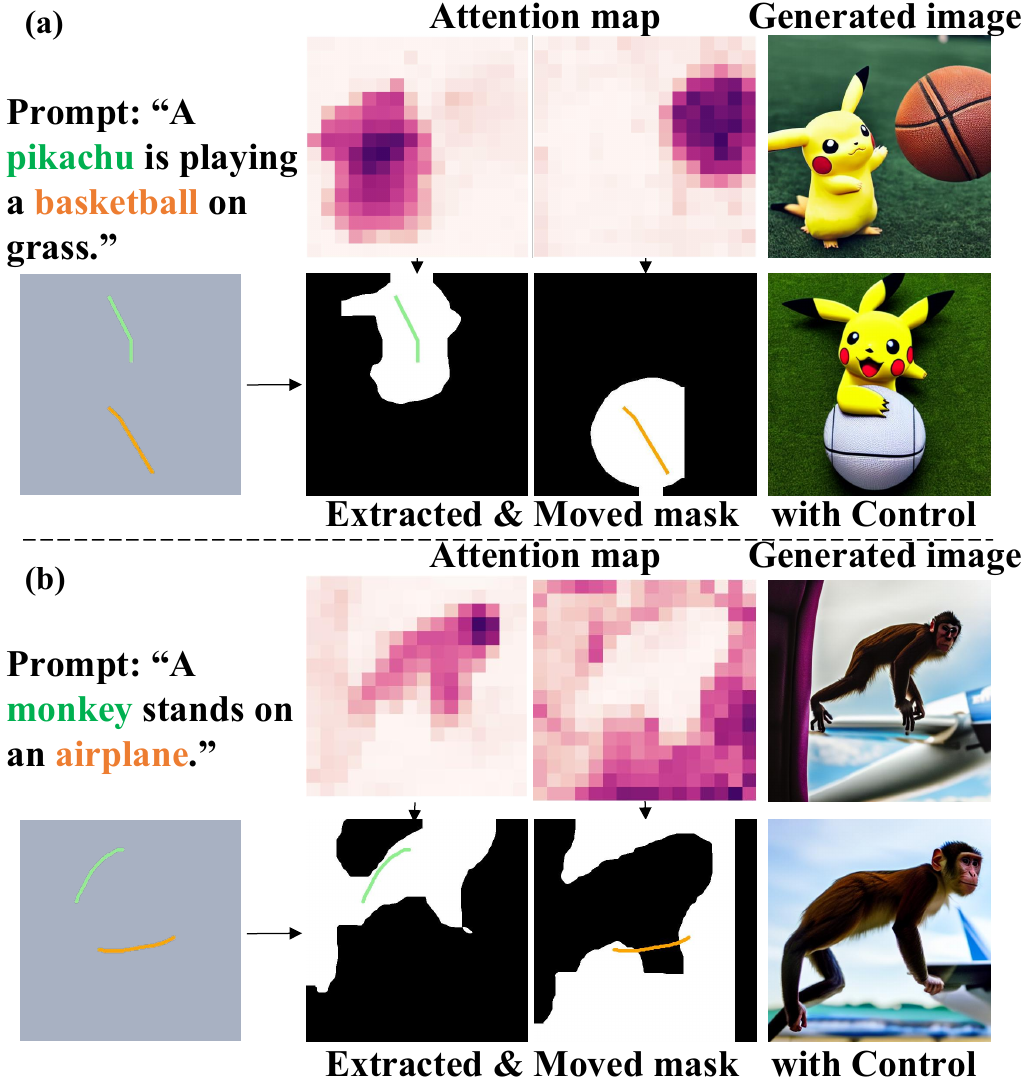}
  \caption{Examples of images generated based on Prior Structure based Guidance. Example (a) shows that over fine mask leads to the generated ``pikachu'' with three ears; and (b) shows that unusable masks are obtained when the pre-trained stable diffusion model generates the poor image. In each example, the top line is the generated image from the pre-trained stable diffusion model with related attention maps, the bottom line is the result based on the trajectory-conditioned Prior Structure based Guidance and related masks through applying the threshold on the attention map and moving to the given trajectory.}
  \label{fig:mask_err}
\end{figure}



\subsection{Is the trajectory similar to scribble?}
\begin{figure*}[t]
  \centering
  \includegraphics[width=0.99\linewidth]{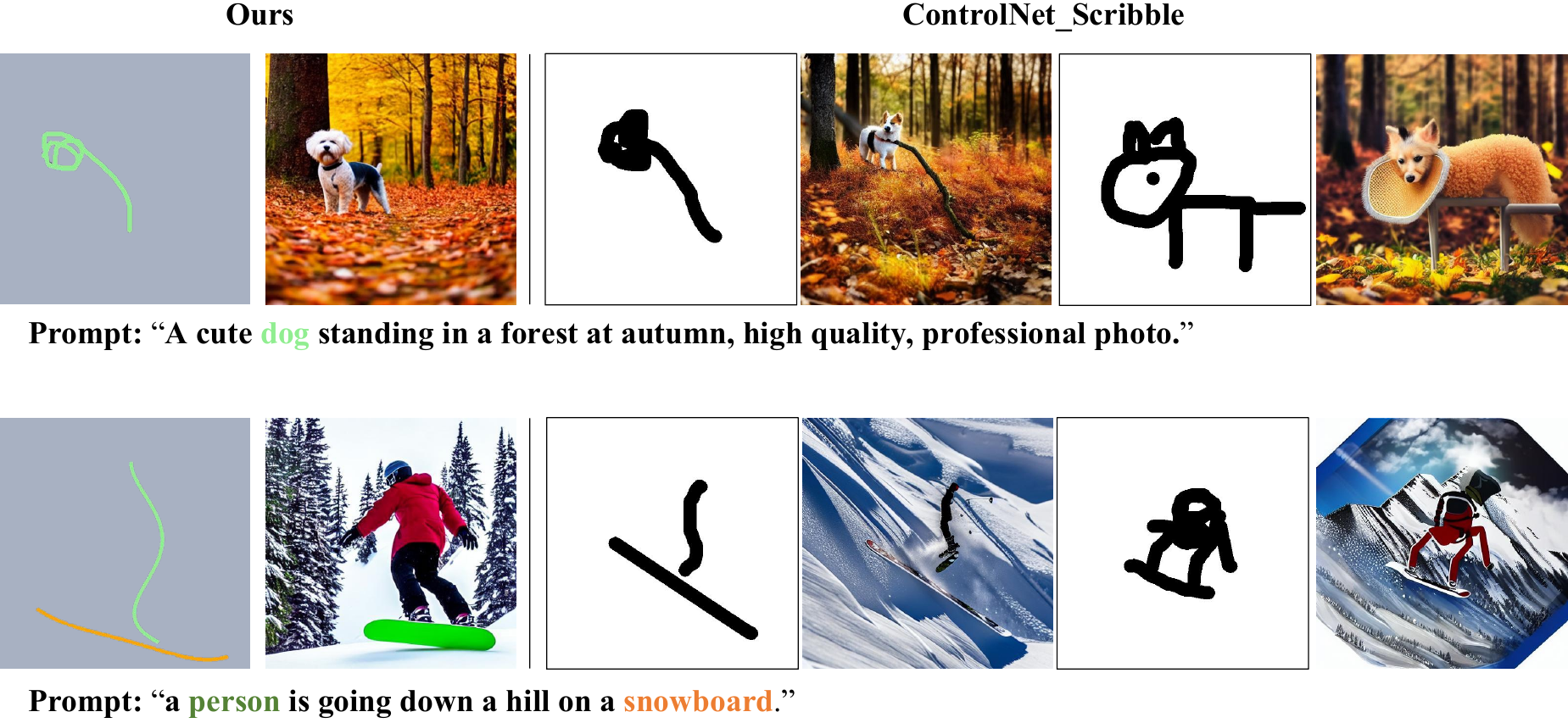}
  \caption{Comparing with ControlNet Scribble(middle and right). We observe that ControlNet with scribble essentially remains a mask-based method, as it cannot be effectively controlled using overly simplistic scribbles.}
  \label{fig:ctrl}
\end{figure*}

We compare our trajectory-based method with ControlNet Scribble, as shown in Figure~\ref{fig:ctrl}, the ControlNet with scribble essentially remains a mask-based method, as it cannot be effectively controlled using overly simplistic scribbles.

\begin{figure*}[t]
  \centering
  \includegraphics[width=0.86\linewidth]{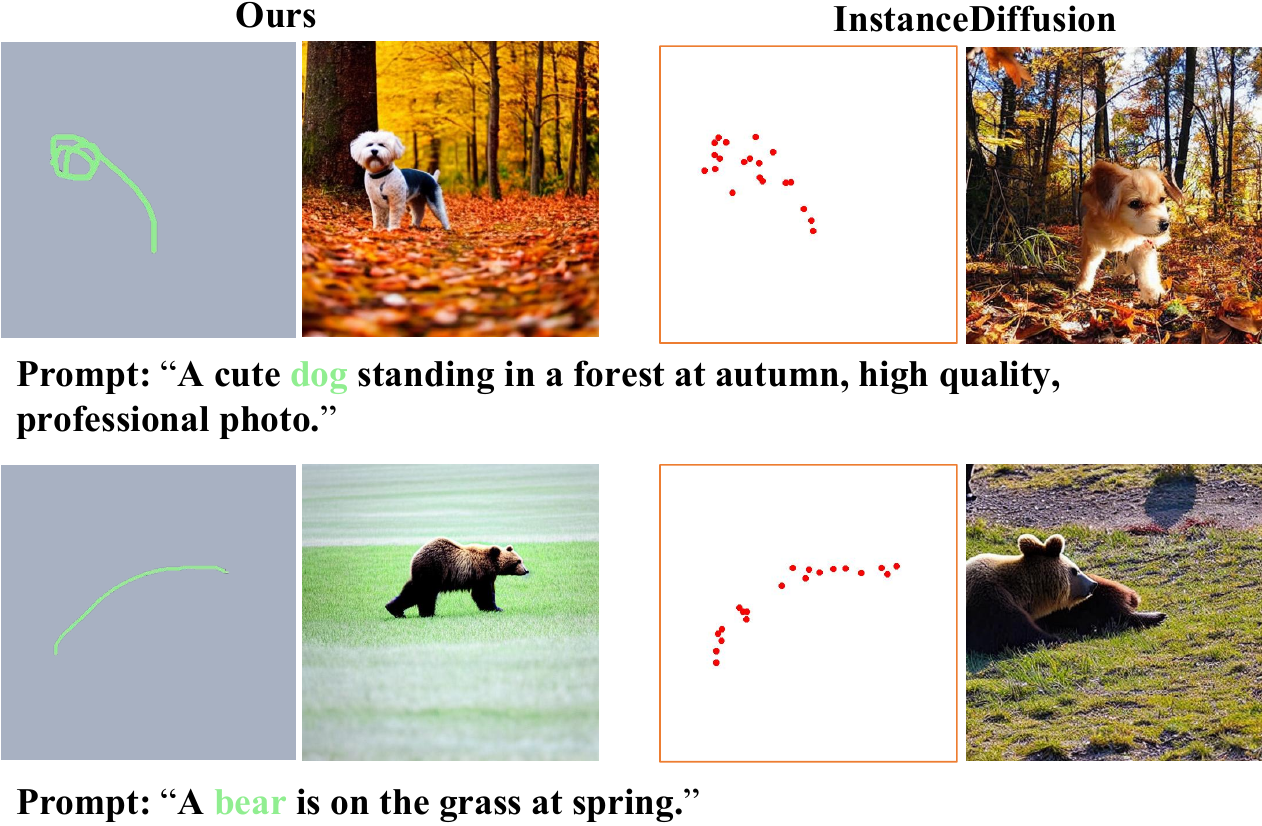}
  \caption{Comparing with InstanceDiffusion Scribble~(right). We observe that InstanceDiffusion with scribble essentially remains a point-based method, it fails to align the generated targets with the provided scribble points.}
  \label{fig:insdiff}
\end{figure*}

We also compare the recently proposed InstanceDiffusion. InstanceDiffusion is essentially a point-based method, and we observe that its scribble input supports a maximum of 20 points. Therefore, we randomly sample 20 points along the trajectory to serve as its input. As shown in Figure~\ref{fig:insdiff}, InstanceDiffusion generates targets that are not aligned with the given scribble points.

\subsection{The Effect of Different Random Seeds}
\begin{figure*}[t]
  \centering
  \includegraphics[width=0.99\linewidth]{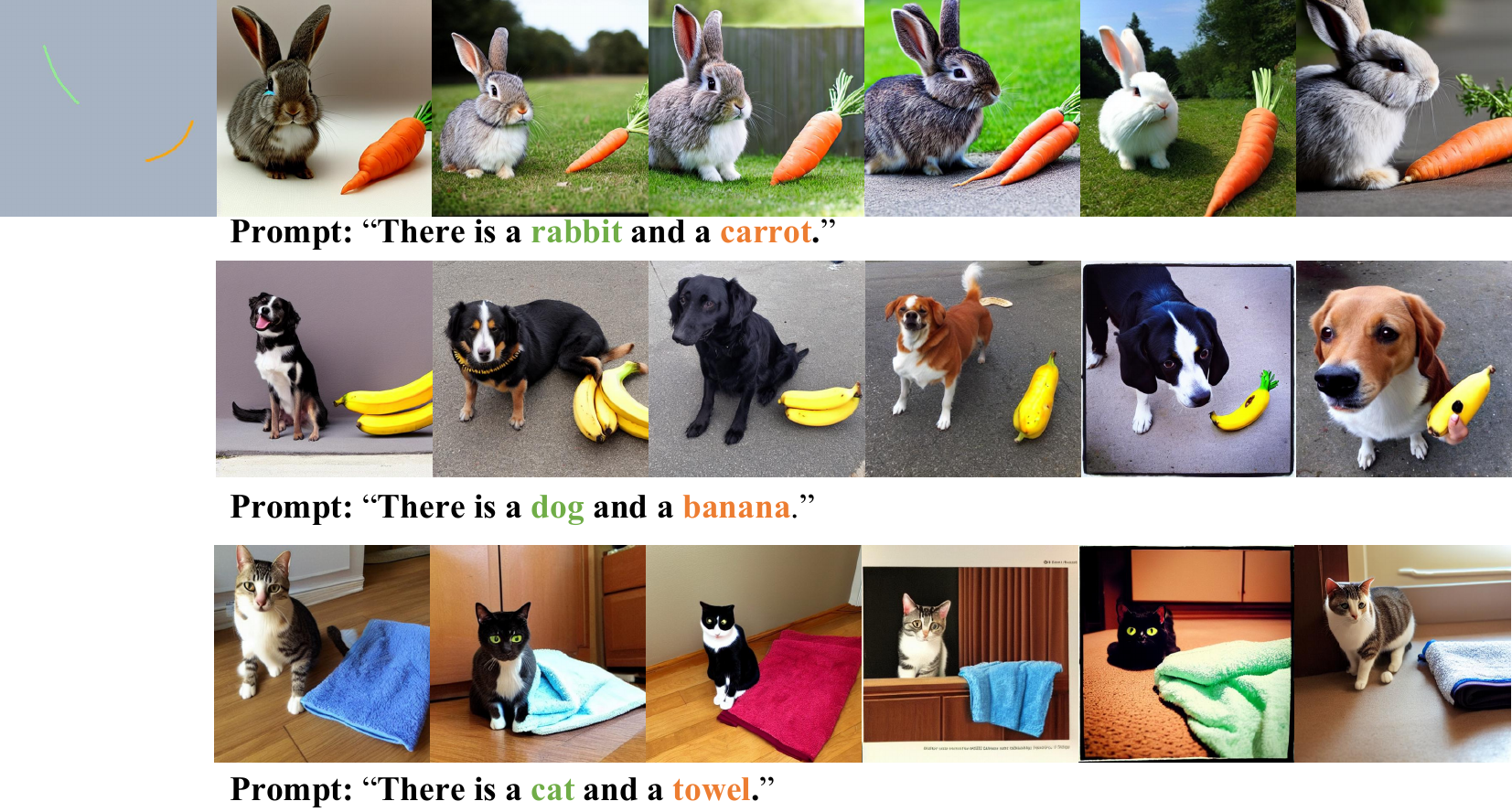}
  \caption{Examples with different random seeds. Our method can reliably achieve control over the targets.}
  \label{fig:seed}
\end{figure*}

We validate the impact of different random seeds on the outcomes of our method, as shown in Figure~\ref{fig:seed}, our method can reliably achieve control over the targets.

\begin{figure*}[t]
  \centering
  \includegraphics[width=0.98\linewidth]{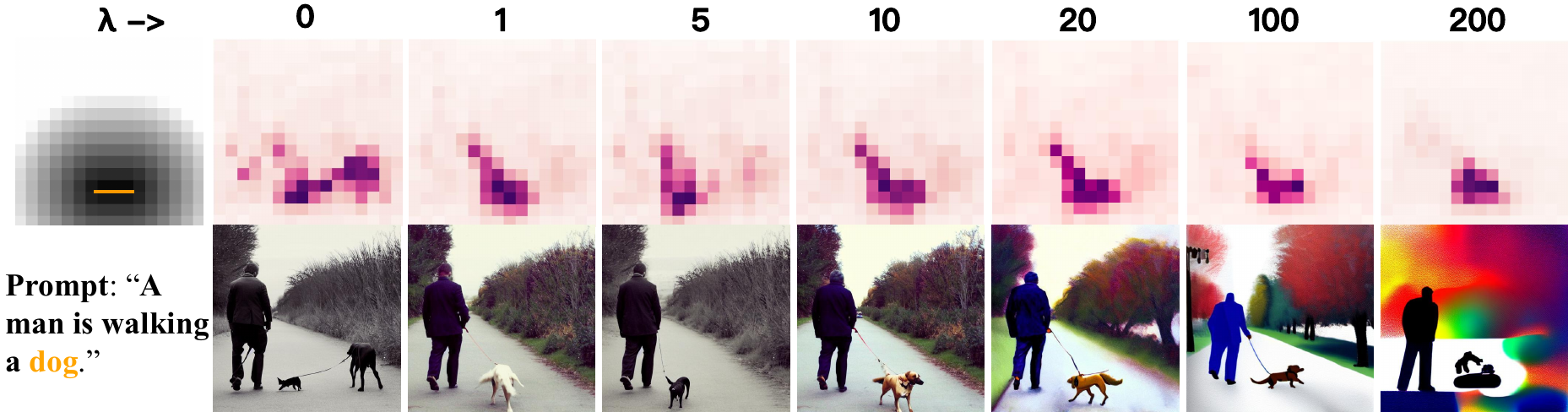}
  \caption{Qualitative analysis the effect of the different $\lambda$. The values in the range of 5-10 yielded the best results.}
  \label{fig:hyper}
\end{figure*}


\end{document}